\documentclass{article}

\usepackage{url}
\usepackage{graphicx}
\usepackage{authblk}

\begin{document}

\title{Convolutional Neural Networks for Aircraft\\
Noise Monitoring}

\author[1]{\normalsize Nicholas Heller}
\author[2]{\normalsize Derek Anderson}
\author[2]{\normalsize Matt Baker}
\author[2]{\normalsize Brad Juffer}
\author[1]{\normalsize \\Nikolaos Papanikolopoulos}
\affil[1]{\small Computer Science and Engineering, University of Minnesota}
\affil[2]{\small Metropolitan Airports Commission}

\date{}

\renewcommand\Authands{ and }

\maketitle

\begin{abstract}
Air travel is one of the fastest growing modes of transportation, however, the effects of aircraft noise on populations surrounding airports is hindering its growth. In an effort to study and ultimately mitigate the impact that this noise has, many airports continuously monitor the aircraft noise in their surrounding communities. Noise monitoring and analysis is complicated by the fact that aircraft are not the only source of noise. In this work, we show that a Convolutional Neural Network is well-suited for the task of identifying noise events which are not caused by aircraft. Our system achieves an accuracy of 0.970 when trained on 900 manually labeled noise events. Our training data and a TensorFlow implementation of our model are available at \url{https://github.com/neheller/aircraftnoise}.
\end{abstract}

\section{Introduction}
A number of studies have been conducted to evaluate aircraft noise and impacts to individuals' health \cite{healtheffectsreview, healtheffectsschoolchildren, healtheffectssleep, healtheffectshypertension}, and on the economy of affected areas \cite{economiceffecturbanproperty, economiceffectresidentialproperty, economiceffectvaluation}. Potential effects are especially important near airports due to high densities of aircraft flying at low altitudes. It has become commonplace for major airports to have
hundreds of arrivals or departures per day, some producing noise events in excess of 80dB SPL in nearby populated areas. 

For these reasons, considerable investment has been made by engine and airframe manufacturers and aerospace designers to reduce the amount of noise aircraft produce. Additionally, airports have invested in mitigating noise through land use planning and management activities, like sound insulating homes and schools, and many airports collaborate with Federal Aviation Administration officials to reduce aircraft flights over noise-sensitive areas \cite{noisemitigationrunways, noisemitigationatc, noisemitigationjatm, noisemitigationautomation}. In planning and evaluating the effectiveness of these noise abatement strategies, efficient monitoring of the aircraft noise affecting areas around major airports is crucial \cite{noisemitigationmeasurement}. 

\begin{figure}[!t]
\centering
\includegraphics[width=3.3in]{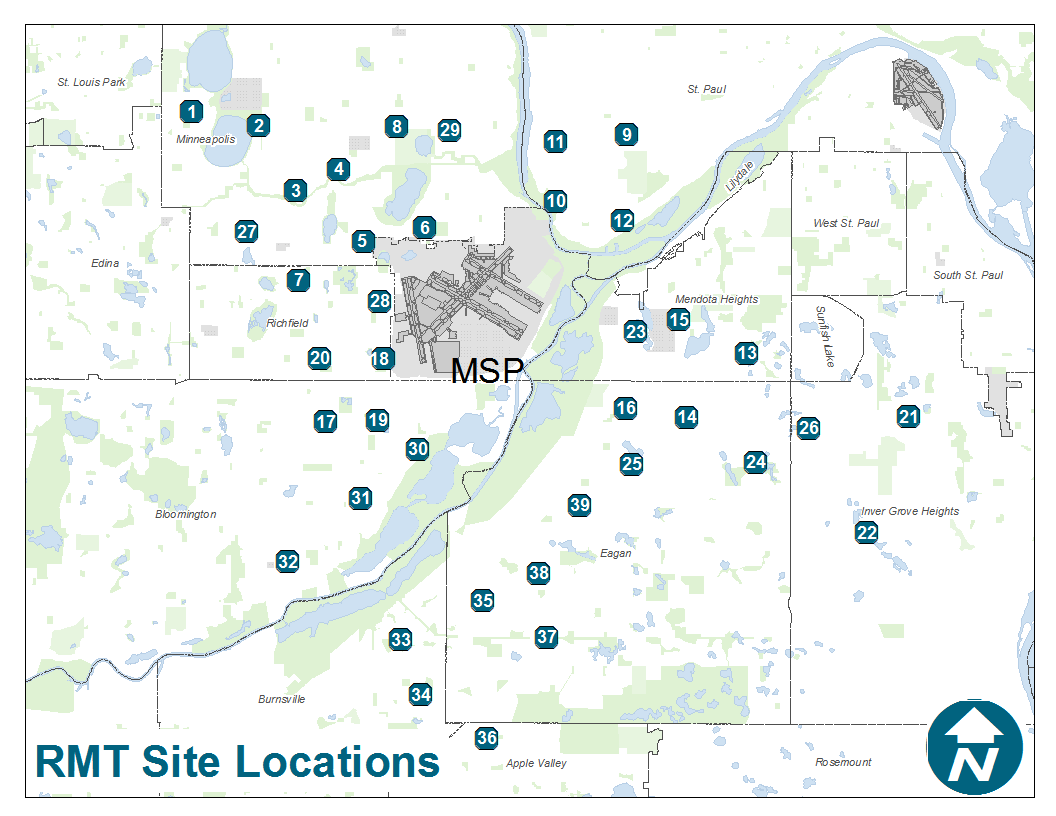}
\caption{A map of the 39 remote noise monitoring sites surrounding the Minneapolis St. Paul International Airport}
\label{fig_monitoringstations}
\end{figure}

The Metropolitan Airports Commission (MAC) has been operating an extensive Noise Monitoring System (MACNOMS) since 1992 that currently includes 39 monitoring stations spread throughout mostly residential neighborhoods surrounding the Minneapolis--St. Paul International Airport (Fig. \ref{fig_monitoringstations}). Each monitoring station is equipped with a Larson Davis 831 Sound Analyzer \& Outdoor Microphone System and monitoring is performed continuously. As is common practice, the MAC stores segments of the data stream -- termed \textit{events} -- where the sound pressure exceeds 65 dBA and maintains 63 dBA or greater for at least 8 seconds.

Not all noise events are caused by aircraft. For instance, lawnmowers, snowplows, and thunder are non-aircraft events commonly collected by MACNOMS. 

It is important for data integrity and reporting accuracy that the MAC is able to reliably identify these non-aircraft events and remove them from consideration. Until recently, the MAC classified a limited number of noise events manually by listening to the audio recordings -- a tedious and inefficient process.

Another consideration in this problem is the expense involved in equipping these remote monitoring stations with high-speed communication infrastructure. Therefore, we set out to perform this classification on as small of a representation of each event as possible, so that all events could be easily communicated and stored centrally in order to maintain a historical record. 

In this work, we present an automated machine-learning based method for classifying the noise events using the one-second LAeq of 1/3 octave frequencies. Our method is based on recent advances in image classification, and is continually improving through human feedback.

\section{Related Work}
Due to the significant effects of aircraft noise near airports, many major airports continuously monitor aircraft noise \cite{afcommunityannoyance, communityannoyancekorea, communityannoyanceamsterdam, communityannoyancetaiwan}, and a number of recent works have offered approaches for classifying collected noise events. 

In \cite{realtimelikeness} the authors implemented a classification scheme based on analyzing the spectral content of the raw audio data. They first used windowing and then transformed into frequency space using the Fast Fourier Transform, ultimately extracting the Mel Frequency Cepstral Coefficients (MFCC) as features. The authors then fit a mixture of Gaussian distributions to the features from each class and used Bayes rule for prediction. Their data-set consisted of over 2700 labeled events from the surroundings of the Madrid Barajas International Airport. 

Similarly, in \cite{featureextraction} the MFCC coefficients were computed along with some hand-selected features on the 1/24 octave data for each event and fed both to separate feed-forward Artificial Neural Networks (ANNs) for classification. In \cite{patternrecognition} the authors trained an ANN on the spectral forms (after some smoothing) of 24 second windows containing noise events to perform this classification. 

In \cite{realtimediscrimination} the authors used a sparse array of 12 microphones to provide more attributes about sound events in order to perform the classification. This sort of approach has a sound theoretical basis, but it comes at a larger equipment and data volume overhead.

Recently, significant advances have been made in the area of learning representations for data jointly with classification training for classification. In particular, Convolutional Neural Networks (CNNs) have surpassed the previous state of the art results for nearly every task in the field of image recognition \cite{alexnet}. To the authors’ knowledge, these methods have not yet been applied to this problem. 

The main contribution of this work is to show that this problem can be re-framed as an image recognition task in which state of the art results can be achieved by training a CNN on low octave resolution at a low sampling frequency.

\section{Methods}

\subsection{Data Collection}
\begin{figure}[!t]
\centering
\includegraphics[width=3in]{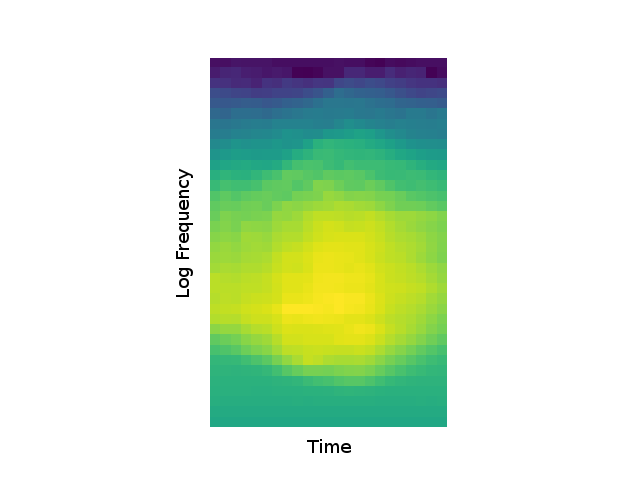}
\caption{An example of the LAeq collected for each 1/3 octave for an event produced by an aircraft. Each row represents a frequency and each column represents a point in time.}
\label{fig_visualization1}
\end{figure}

\begin{figure}[!t]
\centering
\includegraphics[width=3.1in]{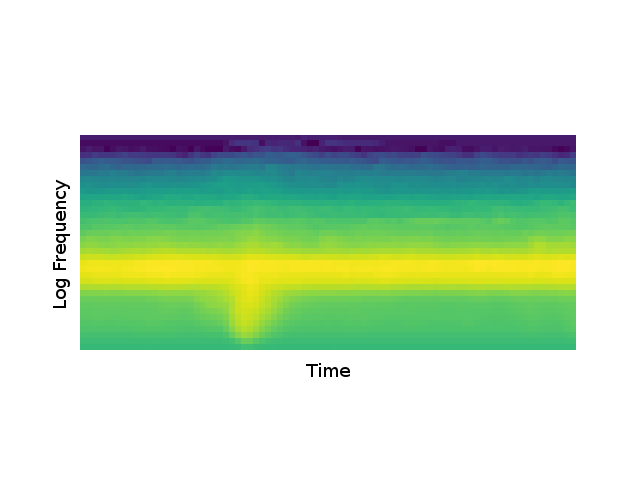}
\caption{An example of the LAeq collected for each 1/3 octave for an event that was not produced by an aircraft. Each row represents a frequency and each column represents a point in time.}
\label{fig_visualization3}
\end{figure}

\begin{figure}[!t]
\centering
\includegraphics[width=3in]{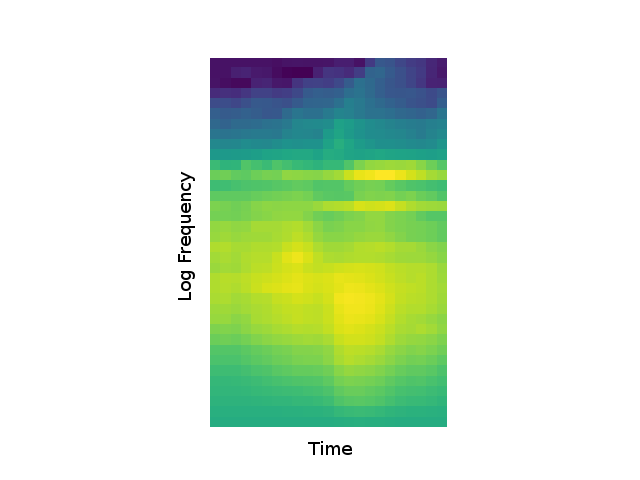}
\caption{An example of the LAeq collected for each 1/3 octave for an event that was not produced by an aircraft. Each row represents a frequency and each column represents a point in time.}
\label{fig_visualization4}
\end{figure}

The MAC is currently collecting and storing the one-second LAeq at every 1/3 octave from 6.3 Hz to 20,000 Hz from every noise event detected by each of their 39 monitoring stations. 900 of these events were selected at random and manually labelled with their true source. Manual labeling was performed by listening to audio recordings of the events. Visualizations of the octave data for a few of these events along with their ground truth label can be seen in Figs. \ref{fig_visualization1}-\ref{fig_visualization4}.

\subsection{Preprocessing}
\begin{figure}[!t]
\centering
\includegraphics[width=3.3in]{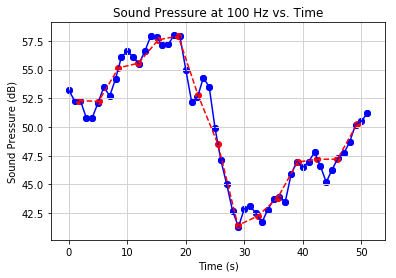}
\caption{An illustration of the interpolation procedure used on each 1/3 octave to map events to the same dimensionality}
\label{fig_interp}
\end{figure}

In order to use a Convolutional Neural Network on this
task, we first had to transform the variable length events into some object with fixed size, which would become the input to the model. To this end, we used linear interpolation to sample uniformly from each of the 36 1/3 octave time series for each event. An example of our sampling for a single octave of one event is shown in Fig. \ref{fig_interp}. In order to preserve any predictive information that the event duration added, we store this as an added feature for classification, which enters our network at the first dense layer.

Once this interpolation was performed and the duration was extracted, we normalize the octave data to have zero mean and unit standard deviation for numerical stability and reducing covariate shift \cite{batchnorm}.

\subsection{Classification}
With natural images, a Convolutional Neural Network makes the assumption that the relationships between pixels near each other in the image are more important to the discrimination task than pixels far from each other. We hypothesized that this assumption holds too in the images we've constructed for this task -- that is, that the relationship between sound pressures near each other in time-frequency space are more important than those far from each other in this space. Some qualitative evidence for this is the fact that the images generated (see Figs. \ref{fig_visualization1}-\ref{fig_visualization4}) appear distinct to humans, whose visual processes also exploit this assumption. 

\begin{figure}[!t]
\centering
\includegraphics[width=2.7in]{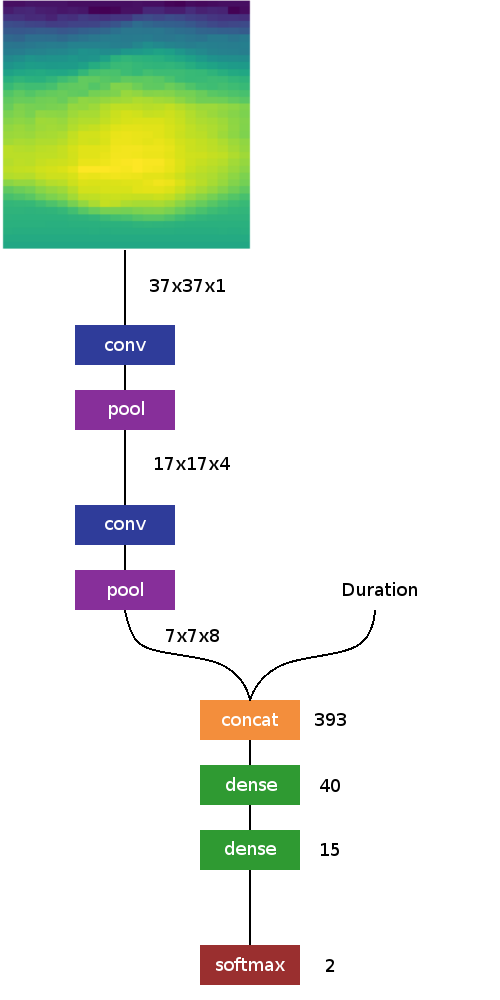}
\caption{A diagram showing the architecture of the CNN. The convolutions are both using 3x3 kernels with unit stride and no padding. The pooling layers are both 2x2 max pooling, also unpadded. A rectified linear unit activation function along with dropout was applied after each convolution and dense layer. There are 36 1/3 octaves between 6.3 and 20,000 Hz, the 37th row is the overall LAeq. Each column represents an interpolated LAeq in time. The number of samples (37) was optimized as a hyperparameter.}
\label{fig_architecture}
\end{figure}

We implemented our Convolutional Neural Network using TensorFlow's Python API \cite{TensorFlow}. Our model is similar to LeNet-5 \cite{lenet} in architecture, but diverges from LeNet-5 in its use of modern techniques such as batch normalization after each layer \cite{batchnorm} and dropout regularization \cite{dropout}. A diagram showing our architecture can be seen in Fig. \ref{fig_architecture}. 

We trained our model using the Adam Optimizer with a learning rate of 0.0004, a dropout keep probability of 0.6, and a bootstrap sample of size 2000 for each batch. 10-fold cross-validation was used to measure performance. 

The final model was trained on the entire data-set. This has been deployed in the MAC's production environment, and at time of writing, thousands of events are being fed through the model daily. 

Since 900 events is a relatively small data-set, we hypothesize that the model will continue to improve with the introduction of more labeled data. To this end, we are employing a strategy called active learning via uncertainty sampling. This involves measuring the discrete entropy of the probability distribution produced by the softmax layer of the CNN for each event. If it lies above a certain threshold, we mark it for manual labeling. Periodically, these new labeled events are added to the training set and the model is retrained. The assumption is that these examples will be especially informative during future training rounds, since they likely represent parts of the distribution of events that the model is unfamiliar with. This technique has been experimentally verified in many settings, including convolutional neural networks for image recognition \cite{activecnns}.

\section{Results}
As stated, we evaluated the performance of our model using 10-fold cross validation. For each fold, we trained the network 5 times using different random weight initializations in order to produce as faithful a measurement of expected performance as possible.

When creating the ground truth labels, we chose to equally sample from each class in order to avoid class balancing issues during training. Although it should be noted that this is a highly unbalanced problem, with an order of magnitude more aircraft events collected than community. 

\begin{figure}[!t]
\centering
\includegraphics[width=3.3in]{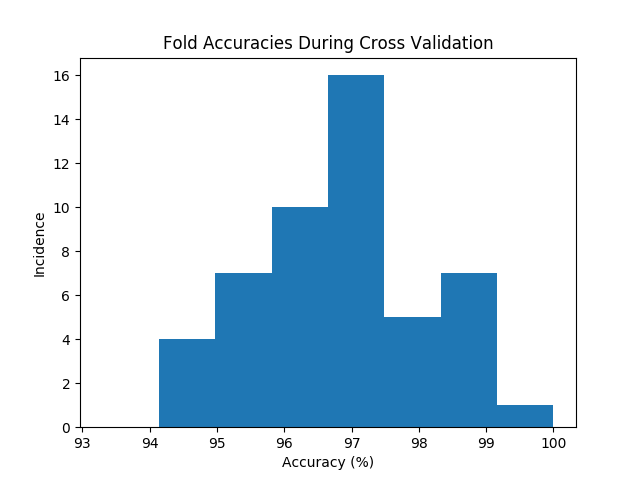}
\caption{A histogram of the final accuracy attained after each training session.}
\label{fig_hist}
\end{figure}

A histogram of the final accuracies of each of our training sessions can be seen in Fig. \ref{fig_hist}. The measurements appear approximately normally distributed. The sample median is 0.970 and the standard deviation is 0.0128. The tightness of this distribution shows the reliability of our method.

\section{Conclusion}
In this work we have shown that with a relatively small training set of 900 labeled events and with relatively low dimensional representations of each event, a convolutional neural network can be trained to achieve state of the art binary classification of aircraft noise events.

In the future, we plan to extend this work to examining the entire time-stream from each monitoring station in order to attempt classification of noises that fall short of the event threshold, but still may have been caused by aircraft. 

In the spirit of reproducible research, we have made our source code and labeled training set available at \url{https://github.com/neheller/aircraftnoise}.

\section*{Acknowledgment}
This material was partially based on work supported by the NSF through grant \#IIP-1439728.

\bibliographystyle{plain}
\bibliography{main}

\end{document}